\documentclass[acmsmall]{acmart}
\AtBeginDocument{%
  }

\setcopyright{acmlicensed}
\copyrightyear{2025}
\acmYear{2025}
\acmDOI{XXXXXXX.XXXXXX.X}

\acmISBN{978-1-4503-XXXX-X/2025/XX}

\usepackage{bm}
\usepackage{subcaption}

\begin{document}

\title{TSCAN: Context-Aware Uplift Modeling via Two-Stage Training for Online Merchant Business Diagnosis}


\author{Hangtao Zhang} 
\email{zht267501@alibaba-inc.com}
\affiliation{%
  \institution{Rajax Network Technology (Taobao Shangou of Alibaba)}
  \city{Hangzhou} 
  \country{China}
}

\author{Zhe Li} 
\authornote{Corresponding author.}
\email{lz171761@alibaba-inc.com}
\affiliation{%
  \institution{Rajax Network Technology (Taobao Shangou of Alibaba)}
  \city{Shanghai} 
  \country{China}
}

\author{Kairui Zhang}
\email{kairui.zhang@alibaba-inc.com}
\affiliation{%
  \institution{Rajax Network Technology (Taobao Shangou of Alibaba)}
  \city{Shanghai} 
  \country{China}
}

\renewcommand{\shortauthors}{Hangtao Zhang, Zhe Li, Kairui Zhang}


\begin{abstract}
Accurate estimation of the Individual Treatment Effect (ITE) is essential for business diagnostics in the online food delivery industry, particularly for assessing the impact of various business strategies, such as inventory management, pricing optimization and online marketing campaigns. 
A primary challenge in ITE estimation lies in sample selection bias. Conventional approaches utilize treatment regularization techniques such as Integral Probability Metrics (IPM), re-weighting, and propensity score modeling to mitigate this bias. However, these regularizations may introduce undesirable information loss and limit predictive performance. 
Moreover, treatment effects exhibit substantial heterogeneity across external contextual factors, such as market demand, competitor activity and time dynamics, yet existing methods fail to adequately model the interaction between treatments and context, limiting their causal expressiveness.
To address these issues, we propose TSCAN: a Context-Aware uplift model based on a Two-Stage training approach, comprising CAN-U and CAN-D sub-models. 
In Stage 1, CAN-U generates counterfactual uplift labels while mitigating selection bias through integrated IPM and propensity score regularization.
In Stage 2, CAN-D eliminates these regularizations and leverages an isotonic output layer to directly model uplift effects in a supervised manner. By reinforcing factual outcomes, CAN-D adaptively corrects estimation errors from CAN-U while circumventing the performance degradation induced by bias-mitigation regularizations.
Additionally, both stages incorporate a Context-Aware Attention mechanism that dynamically fuses the embeddings of merchants, treatments and contextual covariates, thereby capturing context-dependent heterogeneity in treatment effects.
We conduct extensive experiments on two real-world datasets to validate the effectiveness of TSCAN. 
Ultimately, the deployment of our model for real-world merchant diagnosis on one of China's largest online food ordering platforms validates its practical utility and impact.

\end{abstract}



\maketitle

\section{Introduction}
\label{sec:Introduction}

In recent years, the e-commerce sector, particularly the food ordering industry, has seen rapid growth, with China emerging as the largest market, boasting a size of \$40.2 billion in 2024~\cite{1}. This growth has prompted an increasing number of independent merchants to shift towards online sales~\cite{2}. 
However, many of these merchants lack sufficient experience in online operations, which hinders their effective utilization of complex management workflows and diverse marketing tools~\cite{3}. To assist these merchants in accurately identifying business issues and delivering personalized solutions, it is crucial to assess the impact of each diagnostic on their operations. This problem differs from traditional supervised learning in that it requires causal inference rather than mere association modeling. In real-world scenarios, we typically only observe a merchant's response to a specific marketing strategy (i.e., whether they participated in a particular marketing activity or not), but rarely observe their performance under both participating and non-participating strategies simultaneously. To address this problem, researchers have developed methods for modeling individual uplift, known as Individual Treatment Effect (ITE) Estimation~\cite{36,4}. These techniques are primarily used to evaluate the magnitude of the response to an intervention (treatment) across different individuals.
The main ITE estimation methods~\cite{6,7} include: Meta-Learning Methods (e.g., S-Learner~\cite{10}, T-Learner~\cite{10}, X-Learner~\cite{10}), Tree-based Methods (e.g., BART~\cite{11}, CausalForest~\cite{12}), and Deep Learning Methods (e.g., TransTEE~\cite{5}, CEVAE~\cite{8}, CFR-ISW~\cite{CFR_ISW2019}).
Recent research trends indicate that deep learning-based methods are gaining popularity due to their powerful non-linear representation and feature interaction capabilities~\cite{5,7,CFR_ISW2019,14,15}. These deep learning models can be further categorized as follows~\cite{7}:

\begin{itemize}
  \item Balanced Representation Learning (e.g., BNN~\cite{16}, TransTEE).
  \item Covariate Confounding Learning (e.g., CEVAE, Dragonnet~\cite{9}).
  \item Generative Adversarial Network (GAN)-based models (e.g., CEGAN~\cite{17}, GANITE~\cite{18}, SCIGAN~\cite{19}).
\end{itemize}

These methods employ a variety of techniques to mitigate sample selection bias and the adverse effects of confounding factors. For instance, balanced representation learning typically leverages the Integral Probability Metric (IPM)~\cite{mmd2012} as a regularization term to minimize the distributional discrepancy between representations under control ($t=0$) and treatment ($t=1$) conditions, i.e., between $p(\Phi(x)|t=0)$ and $p(\Phi(x)|t=1)$~\cite{20,21}. Additionally, CFR-ISW~\cite{CFR_ISW2019} enhances this approach by incorporating a re-weighting strategy, in which the sample weight is related to the propensity score. Methods based on covariate confounding learning primarily encode both observed and unobserved confounders using approaches such as autoencoders or propensity score prediction networks, thereby eliminating the influence of these confounding factors~\cite{23,24,25}. 

Despite these advances, several limitations persist in current ITE estimation frameworks. Many branch-structured neural architectures and meta-learning algorithms remain restricted to discrete treatments and cannot naturally accommodate continuous treatment spaces. Furthermore, strategies designed to correct for selection bias often introduce auxiliary estimation errors~\cite{26}. For example, overly stringent balancing constraints may inadvertently discard outcome-predictive features, degrading model performance~\cite{26,27}. Similarly, reweighting based on propensity scores can distort the empirical data distribution, disproportionately amplifying the influence of rare or long-tail samples~\cite{28}. Incorporating a propensity score prediction task will also affect the outcome prediction~\cite{CFR_ISW2019}, as evidenced by empirical observations that ablating such regularization components often improves accuracy on observed outcomes under identical training conditions.

Beyond bias correction, a critical gap lies in the inadequate integration of contextual information into treatment effect modeling. In dynamic environments such as online marketing, the efficacy of a given intervention (treatment, e.g., a promotional subsidy or ad bid) is highly context-dependent. For instance, consumer responsiveness to merchant subsidies tends to diminish in high-demand markets, whereas sensitivity increases in saturated or competitive markets with abundant alternatives. Similarly, the impact of a fixed ad bid varies significantly with the bidding behavior of competitors—a key contextual variable. Such contextual heterogeneity is essential for accurate uplift estimation, yet it remains largely unexploited by existing methods, which typically assume treatment effects are context-invariant. 

In summary, two overarching challenges remain: (1) developing more effective methods to address selection bias while maintaining the predictive performance of the model and the quality of personalized recommendations; (2) accounting for the impact of contextual factors on treatment effects.

The main contributions of this work are:
\begin{enumerate}
    \item We propose TSCAN, a novel two-stage context-aware uplift modeling framework that decouples bias mitigation (Stage 1) from direct uplift prediction (Stage 2), thereby avoiding the performance degradation caused by conventional regularizations.
    
    \item We design a Context-Aware Attention Layer that explicitly models the tripartite interaction among merchant features, treatments, and external contexts, enabling adaptive ITE estimation across diverse operational scenarios.
    
    \item We systematically validate TSCAN through three lenses: (RQ1) \textit{benchmark performance}: showing consistent superiority over seven state-of-the-art baselines on two large-scale real-world datasets; (RQ2) \textit{architectural ablation}: demonstrating the individual value of two-stage training, context-aware attention, and isotonic output; and (RQ3) \textit{real-world impact}: achieving a 0.76\% increase in merchant orders in a live A/B test on one of China's largest food delivery platforms, which demonstrates its practical value.
\end{enumerate}

\section{RELATED WORK}
\label{sec:related_work}

In this section, we provide a concise overview of the primary existing works on uplift models, context-aware treatment effect estimation and feature interaction methods.

\subsection{Uplift Modeling: From Meta-Learners to Deep Causal Representations}

Uplift modeling aims to estimate heterogeneous treatment effects from observational or experimental data, with applications ranging from personalized medicine~\cite{30} to digital marketing~\cite{31}. Early approaches adopted meta-learning frameworks (e.g., S-, T-, and X-Learners~\cite{10}), which repurpose standard regressors but often fail to capture complex treatment-covariate interactions. Tree-based methods such as Bayesian Additive Regression Trees (BART)~\cite{11} and Causal Forest~\cite{12} improve interpretability and perform well under low-dimensional settings. However, they struggle to effectively model the high-dimensional and sparse features typical of online platforms.

Recent advances leverage deep neural networks to learn balanced representations that mitigate confounding bias~\cite{37,5,7,14,32}. Notable examples include TarNet~\cite{20} and CFR~\cite{CFR_ISW2019}, which use IPM such as Maximum Mean Discrepancy (MMD) to align treatment and control group distributions in latent space. Dragonnet~\cite{9} and CEVAE~\cite{8} integrate propensity score estimation to adjust for selection bias, while GAN-based models (e.g., GANITE~\cite{18}) simulate counterfactual outcomes. However, as highlighted in~\cite{26,27}, strict enforcement of distributional balance can discard outcome-predictive information, leading to degraded practical performance—a key limitation our two-stage design seeks to overcome.


\subsection{Context-Aware Treatment Effect Estimation}

Recent studies increasingly recognize that treatment effects are not static but modulated by external contexts. Huang et al.~\cite{huang2024entire} propose ECUP, a context-enhanced uplift model that leverages user behavior chains and campaign timing to capture phase-dependent treatment efficacy in marketing. Separately, UMLC~\cite{sun2025robust} addresses robustness in real-time interventions by grouping large-scale contextual features (e.g., merchant, time and location) via response-guided clustering, enabling stable ITE estimation across volatile environments. Though not causal, Afzal et al.~\cite{afzal2024multi} demonstrate in recommendation systems that fusing multi-dimensional contexts—temporal, geographic, and social—through deep interaction layers significantly improves context-dependent response modeling, a principle highly relevant to uplift.

These works underscore a critical insight: \textit{the same treatment can yield divergent outcomes under different contextual conditions}. For example, a merchant discount may significantly increase order volume during off-peak hours but have negligible effect during lunchtime peak periods due to demand saturation. Despite this, most existing uplift models—including TransTEE~\cite{5}, EFIN~\cite{14}, and CFR-ISW~\cite{CFR_ISW2019}—either treat context as ordinary covariates or ignore it entirely, failing to model the tripartite interaction among \textit{merchant}, \textit{treatment} and \textit{context}.

Our work bridges this gap by explicitly designing a \textit{Context-Aware Attention Layer} that dynamically reweights merchant and treatment representations based on contextual embeddings. Through this, we enable end-to-end interaction within a unified architecture.

\subsection{Feature Interaction in Causal Models}

Accurately capturing how treatments interact with individual characteristics is essential for modeling heterogeneous effects. A growing line of work focuses explicitly on feature interaction in causal models. For instance, TransTEE~\cite{5} encodes continuous treatments into semantic embeddings and fuses them with covariates through attention mechanisms to estimate dose-response functions. EFIN~\cite{14} proposes an Explicit Feature Interaction-aware Uplift Network that employs a treatment-gated module to dynamically adjust the importance of customer and product features based on the treatment type, thereby modeling how different offers motivate distinct user segments. MTMT~\cite{wei2024mtmt} introduces a treatment–user feature interaction module to model correlations between treatments and user features. However, these approaches primarily model direct pairwise interactions between treatment and covariates, assuming that the interaction pattern is fixed across all samples. In this work, we combine a \textit{Treatment-Aware Attention Network} with a \textit{Context-Aware Attention Layer} that explicitly models the dynamic treatment-feature interaction.

In summary, while prior work has made significant strides in representation balancing and treatment-aware interaction, none simultaneously addresses (1) the trade-off between bias correction and predictive fidelity, (2) the dynamic modulation of treatment effects by external context. The proposed TSCAN framework is designed specifically to resolve these dual challenges.

\section{PRELIMINARIES}
In the scenario of estimating treatment effects for business diagnosis in the online food ordering industry, 
our goal is to estimate the ITE using the observed data $D = {\{(X_i , t_i , y_i)\}}_{i=1}^N$, where $X_i$, $t_i$, and $y_i$ represent the merchant features, treatment feature, and outcome value, respectively. 
The treatment variable $t_i$ can be binary, indicating whether a merchant has initiated a marketing activity (i.e., $t_i \in \{0,1\}$), or continuous, such as the number of customer reviews (i.e., $t_i \in \mathbb{R}$). The outcome variable $y_i$ is continuous, representing the merchant's order count or revenue (i.e., $y_i \in \mathbb{R}$). The sample size is denoted by $N$. The potential outcome of the $i$-th instance under treatment value $k$ is denoted as $y_i(t_i=k)$, and the conditional probability of assigning treatment $k$ given features $X_i$ is expressed as $\pi(X_i,k)=P(t_i=k \mid X_i)$, commonly referred to as the propensity score. In observational data, we can typically observe the outcome values under a specific treatment rather than under all possible treatments. This limitation distinguishes uplift modeling from traditional supervised learning. Uplift modeling seeks to accurately estimate the expected outcome for each instance across different treatments (under ignorability assumption):
\begin{align}
    \tau(X_i,k_1,k_0) &= E[y_i(k_1) - y_i(k_0) \mid X_i] \\
    &= E[y_i \mid t_i=k_1, X_i] - E[y_i \mid t_i=k_0, X_i]
\end{align}

\section{THE PROPOSED METHOD}
\subsection{Design Rationale}
To address the dual challenges identified in Section~\ref{sec:Introduction}: (1) The performance degradation caused by conventional treatment regularizations (e.g., IPM~\cite{mmd2012}, propensity score prediction~\cite{Propensity_score2023}), (2) The underutilization of contextual features in modeling treatment effect heterogeneity, this paper introduces \textbf{TSCAN} (\textbf{T}wo-\textbf{S}tage \textbf{C}ontext-\textbf{A}ware uplift \textbf{N}etwork), a novel uplift modeling framework comprising two sub-models: \textbf{CAN-U} and \textbf{CAN-D}.

The core insight of TSCAN is to decouple bias mitigation from direct uplift prediction. In Stage 1, CAN-U is trained with IPM and propensity score regularization to generate high-quality pseudo-uplift labels while reducing selection bias. In Stage 2, CAN-D leverages these labels to perform supervised uplift learning—but crucially \textit{without} the regularizations that compromise predictive accuracy. Instead, CAN-D employs an \textit{isotonic output layer}~\cite{DeepIsotonic2024} to directly model uplift in an interpretable manner. This two-stage design allows TSCAN to enjoy the benefits of bias correction while avoiding its pitfalls.

In the following, we first present the TSCAN model architecture (including CAN-U and CAN-D, Section~\ref{sec:architecture}), detailing how merchant, treatment, and contextual features are fused through attention mechanisms. We then describe the two-stage training procedure (Section~\ref{sec:training}), which enables TSCAN to balance causal robustness and predictive fidelity.

\subsection{Model Architecture of TSCAN}\label{sec:architecture}

\begin{figure}[h]
  \centering
  \includegraphics[width=\textwidth]{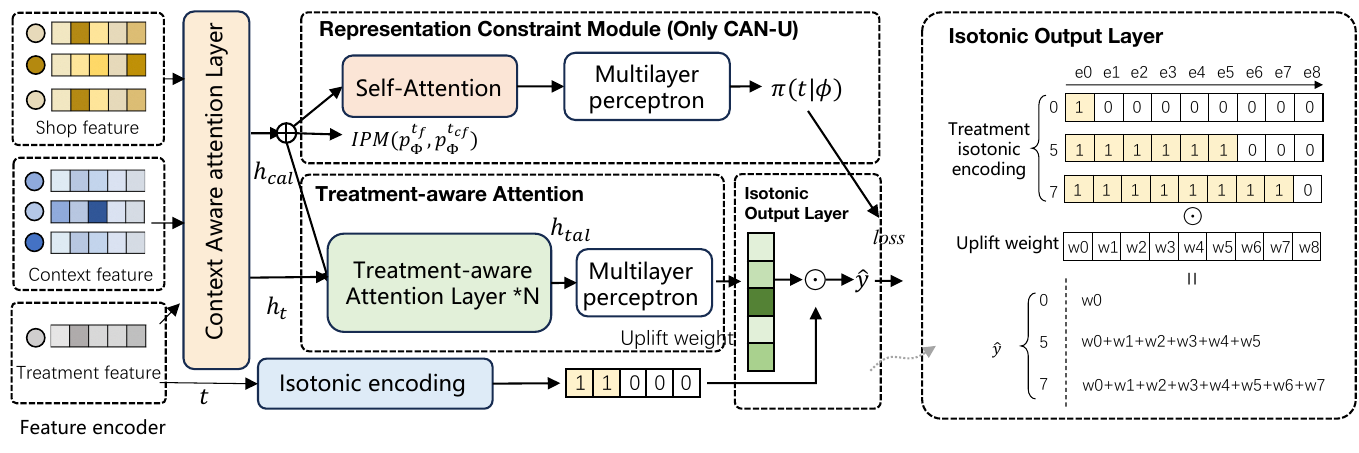}
  \caption{The network architecture of CAN-U and CAN-D.}
  \label{fig:TSCAN}
\end{figure}

\subsubsection{Overall Architecture}
Both CAN-U and CAN-D adopt the same backbone architecture, shown in Figure~\ref{fig:TSCAN}. The only structural difference lies in the presence of the \textit{Representation Constraint Module} (active only in CAN-U) and the use of the \textit{Isotonic Output Layer} (in both models, but trained differently). The data flow is as follows:
\begin{itemize}
\item \textbf{Feature Encoder}: Converts raw features into dense embeddings.

\item \textbf{Context-Aware Attention Layer}: Fuses merchant and treatment embeddings with contextual embeddings to produce context-adaptive representations. This module directly addresses the limitation of existing models in their inadequate incorporation of relevant contextual factors~\cite{huang2024entire,sun2025robust}.

\item \textbf{Representation Constraint Module} (CAN-U only): For CAN-U, this module applies IPM loss and propensity score prediction to mitigate selection bias during training.

\item \textbf{Treatment-Aware Attention Network}: Further refines the merchant representation by conditioning on the treatment. This final representation encapsulates the context-dependent relationship between the merchant and the treatment.

\item \textbf{Isotonic Output Layer}: Directly models uplift effects in a regularization-free and supervised manner.
\end{itemize}

We now detail each component.

\subsubsection{Feature Encoder}
This module converts raw merchant, contextual, and treatment features into dense and comparable embeddings. Each merchant instance is represented by a tuple $(X, t, C, y)$, where $X$ denotes merchant-specific attributes (e.g., rating and operating hours), $t$ is the treatment (binary or continuous), $C$ is the external context (e.g., time-of-day, district type and supply–demand ratio), and $y$ is the observed outcome (e.g., order count). 

Sparse categorical features undergo an embedding table lookup, while continuous features are transformed via a linear layer. The treatment $t$ is encoded in the same manner.

\begin{equation}
    e_k =
    \begin{cases}
        E^{\text{cat}}_k(x^k), & \text{if } k \in \Omega^{\text{cat}} \\
        \mathbf{w}_k x^k + \mathbf{b}_k, & \text{if } k \in \Omega^{\text{num}}
    \end{cases}
\end{equation}
This yields three embedding vectors: merchant embeddings $e_s$, contextual embeddings $e_c$, and treatment embedding $e_t$. These serve as inputs to the subsequent context-aware interaction layers.

\subsubsection{Context-Aware Attention Layer}
\label{sec:context_aware}
This layer dynamically modulates merchant and treatment representations according to the external context, enabling the model to adapt its inference to varying operational environments. For instance, a discount may substantially increase order volume during off-peak hours but exhibit minimal effect during lunchtime peaks owing to saturated demand. To capture such context-dependent heterogeneity, the Context-Aware Attention Layer adaptively reweights merchant and treatment representations based on the current context.

\begin{figure}[h]
    \centering
    \includegraphics[width=0.64\textwidth]{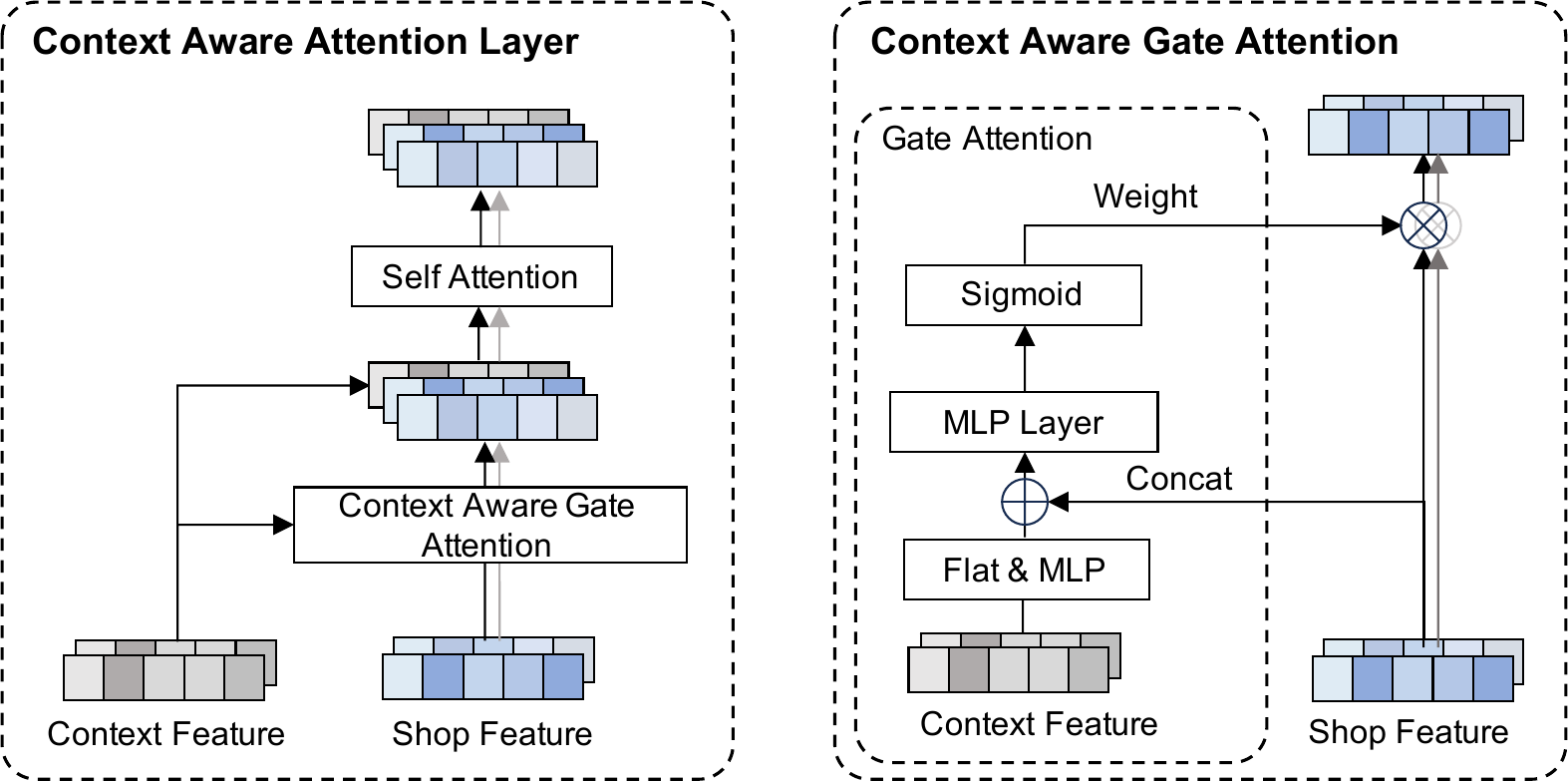}
    \caption{The architecture of Context-Aware Attention Layer and the Context-Aware Gate Attention.}
    \label{fig:context_aware}
\end{figure}

Specifically, for merchant features, the context-aware weight $a_s$ is computed by concatenating $e_s$ with a summarized representation of $e_c$ (via an MLP) and passing it through a sigmoid-activated linear layer. Figure~\ref{fig:context_aware} illustrates this process.

\begin{align}
    a_s &= 1 + \sigma\big( \mathbf{W}_p [\mathbf{e}_s; \text{MLP}(\text{Flat}(\mathbf{e}_c))] + \mathbf{b}_p \big),
\end{align}
where $a_s$ acts as a context-gated attention weight, ensuring that context-relevant merchant features are amplified. $\mathbf{W}_p$ and $\mathbf{b}_p$ are the parameters of the MLP layer, ``;'' denotes concatenation operator, and $\sigma$ refers to the activation function. The adaptive merchant representation is then:
\begin{align}
     \mathbf{h}_s &= a_s \odot \mathbf{e}_s
\end{align}

Finally, a self-attention operation is performed on $[\mathbf{h}_s, \mathbf{e}_c]$ to facilitate richer interactions, producing the final context-aware merchant representation $\mathbf{h}_{\text{cal}}$. A similar gating is applied to $\mathbf{e}_t$:

\begin{align}
    a_t &= 1 + \sigma\big( \mathbf{W}_t [\mathbf{e}_t; \text{MLP}(\text{Flat}(\mathbf{e}_c))] + \mathbf{b}_t \big), \\
    \mathbf{h}_t &= a_t \odot \mathbf{e}_t.
\end{align}

This design ensures that the same merchant–treatment pair yields different latent representations under different contexts, enabling fine-grained ITE estimation.

\subsubsection{Treatment-Aware Attention Network}
This module refines the context-aware merchant representation by explicitly conditioning it on the treatment, thereby modeling the treatment-merchant interaction in a way that is sensitive to the current context.

\begin{figure}[h]
  \centering
  \includegraphics[width=0.64\textwidth]{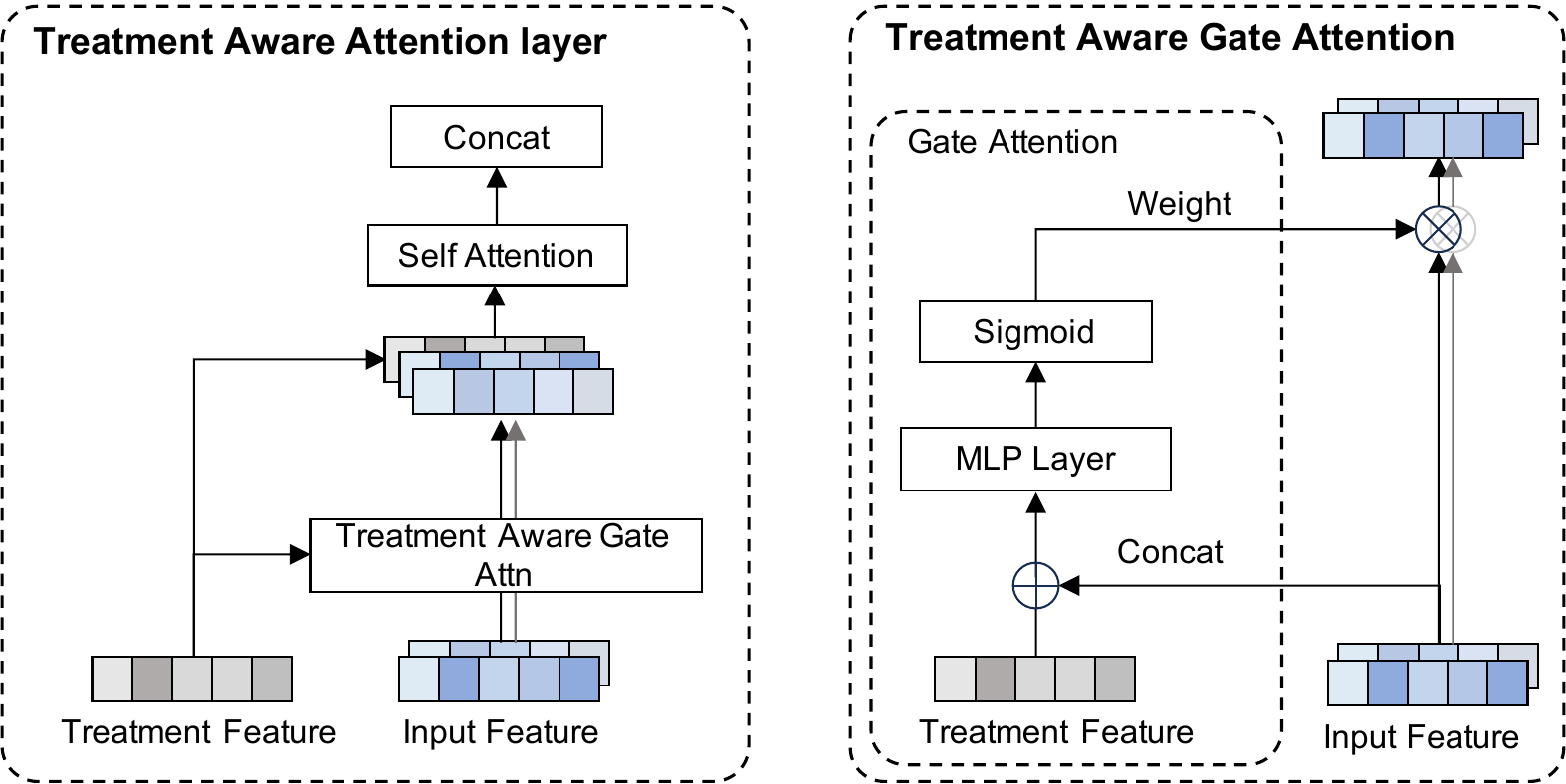}
  \caption{The architecture of Treatment-Aware Attention Layer and the Treatment-Aware Gate Attention.}
  \label{fig:treatment_aware}
\end{figure}

Aiming to model how treatments interact with merchant characteristics, we employ a treatment-gated attention mechanism, but crucially extend it to be context-conditioned. 

Specifically, the merchant representation $\mathbf{h}_{\text{cal}}$ is weighted by a gate $\alpha_{\text{tal}}$ which is a function of both $\mathbf{h}_{\text{cal}}$ and $\mathbf{h}_t$: 
\begin{align}
    \alpha_{\text{tal}} &= 1 + \sigma\big( \mathbf{W}_q [\mathbf{h}_{\text{cal}}; \mathbf{h}_t] + \mathbf{b}_q \big),
\end{align}

where $\mathbf{W}_q$ and $\mathbf{b}_q$ represent the gate attention parameters.
Subsequently, a self-attention operation is performed on $[\alpha_{\text{tal}} \odot \mathbf{h}_{\text{cal}},\mathbf{h}_t]$ to facilitate richer interactions, yielding the final representation:
\begin{align} 
    \mathbf{h}_{\text{tal}} &=\text{Self-Attn}([\alpha_{\text{tal}} \odot \mathbf{h}_{\text{cal}},\mathbf{h}_t]),
\end{align}

where $\mathbf{h}_{\text{tal}}$ captures the contextually adaptive interaction between the merchant and the treatment. This design allows the same merchant-treatment pair to produce different interaction strengths in different environments. Figure~\ref{fig:treatment_aware} illustrates this process.

\subsubsection{Representation Constraint Module (CAN-U Only)}\label{sec:representation_constraint}

This module (exclusively used in CAN-U) injects dual bias-mitigation regularizations: IPM~\cite{mmd2012} and adversarial propensity score estimation~\cite{Propensity_score2023}. Their joint application has been shown to yield more robust distributional alignment than either method in isolation~\cite{CFR_ISW2019,kallus2020deepmatch}.

\begin{itemize}

\item \textbf{IPM}~\cite{mmd2012}: 
IPM (e.g., MMD) is utilized to minimize the distance between the distributions of $h_{cal}^{t=0}$ and $h_{cal}^{t=1}$, aiming to remove confounding by aligning latent representations~\cite{20,mmd2012}. To extend this methodology to continuous treatments, we sort and split the samples into two parts according to treatment values within each batch. Subsequently, we compute the distribution distance between the upper 50\% and lower 50\% of these samples. The IPM loss can be formulated as:
\begin{align}
\mathcal{L}_{\text{IPM}}&=\sup_{||f||_{\mathcal{H}_k} \leq 1 } \left\{ \mathbb{E}_{x \sim p_{t \in T_0}}[f(x)] - \mathbb{E}_{x \sim p_{t \in T_1}}[f(x)] \right\}\\
&=||\mu(p_{t \in T_0})-\mu(p_{t \in T_1})||_{\mathcal{H}_k}
\end{align}

\item \textbf{Propensity Score Prediction}~\cite{Propensity_score2023}:
We estimate the propensity score via an auxiliary network $\pi(\cdot)$. The propensity score regularization accounts for selection bias by ensuring that $f_{\theta}(x, t)$ remains invariant under perturbations, thereby enhancing the robustness of the uplift model~\cite{5}. The propensity score prediction loss can be formulated as:
\begin{equation}
\mathcal{L}_{\pi} = \sum_{i=1}^n \left( t_i - \pi(\hat{t}_i \mid \phi(x_i)) \right)^2
\end{equation}

\end{itemize}

While the integration of IPM and propensity-based regularization improves causal identifiability, it introduces a trade-off: enforcing strict distributional balance may inadvertently suppress outcome-predictive features that are correlated with treatment assignment~\cite{26,27,2026two_stage_repr_balance}. Such regularization-induced information loss can impair the fidelity of outcome prediction. Therefore, rather than using this constrained model as the final estimator, we restrict the Representation Constraint Module to Stage 1 (CAN-U) solely for generating reliable pseudo-uplift labels. The final prediction is delegated to CAN-D in Stage 2, which operates without these regularizations and thus preserves full predictive capacity.

\subsubsection{Isotonic Output Layer}

The Isotonic Output Layer~\cite{DeepIsotonic2024} is introduced to enable direct modeling of uplift effects in a supervised manner, allowing the CAN-D to learn from pseudo-uplift labels generated by CAN-U, while being free from the above regularizations. By explicitly parameterizing the incremental effect of each treatment level, this layer facilitates joint supervision on both factual outcomes and uplift increments, thereby enabling CAN-D to inherit causal knowledge from Stage 1 while adaptively correcting its estimation errors through factual outcome reinforcement (a detailed discussion will be provided in Section~\ref{sec:training}).

Concretely, for a treatment value $t$, $t \in [0,1]$, we discretize it into $M+1$ (M=1 for binary treatment) ordered levels and apply isotonic encoding:
\begin{equation}
\text{IE}(t_i) = [\underbrace{1,\dots,1}_{k+1},\underbrace{0,\dots,0}_{M-k}] ,\quad \text{where } k = \lfloor t_i \cdot M \rfloor
\end{equation}

The model then predicts an uplift weight vector $\mathbf{w}_i = [v_{i,0}, v_{i,1}, \dots, v_{i,M}]^\top$, where each $v_{i,k} \geq 0$ represents the marginal uplift contributed by the $k$-th treatment level. The predicted factual outcome under treatment $t_f$ is:  
\begin{equation}
\hat{y}_{i,f}^{d} = \sum_{k=0}^{k_f} v_{i,k} , \quad \text{with } k_f = \lfloor t_f \cdot M \rfloor
\end{equation}

The uplift from $t_f$ to a counterfactual $t_{cf} > t_f$ is directly computed as the sum over the incremental segment:  
\begin{equation}
\hat{u}_{i}^d = \sum_{k=k_f+1}^{k_{cf}} v_{i,k}
\end{equation}

\begin{figure}[h]
  \centering
  \includegraphics[width=0.49\textwidth]{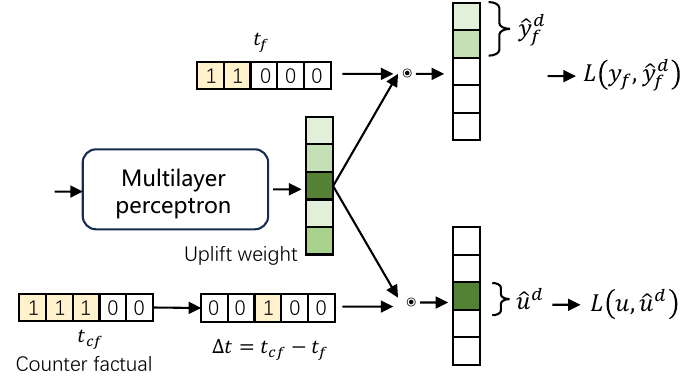}
  \caption{An illustration of prediction process for the factual outcome, counterfactual outcome, and the corresponding uplift.}
  \label{fig:iso}
\end{figure}

This formulation enables CAN-D to be trained with a dual-loss objective (loss of $\hat{y}_f^d$ and $\hat{u}^d$), as demonstrated in Figure~\ref{fig:iso}. By explicitly modeling the incremental effect of treatment levels, the Isotonic Output Layer allows CAN-D to effectively learn from both factual and counterfactual signals, thereby enhancing its uplift estimation accuracy.

This incremental learning strategy decouples factual and counterfactual learning, reducing error propagation—since $y_f$ is observed, its prediction is well-constrained, while $\tilde{u}$ only needs to model the difference, which is typically lower-variance than the full counterfactual outcome~\cite{zhang2025doubly,dr2020incremental}.


\subsection{Two-Stage Training Process}
\label{sec:training}

\begin{figure}[h]
  \centering
  \includegraphics[width=\linewidth]{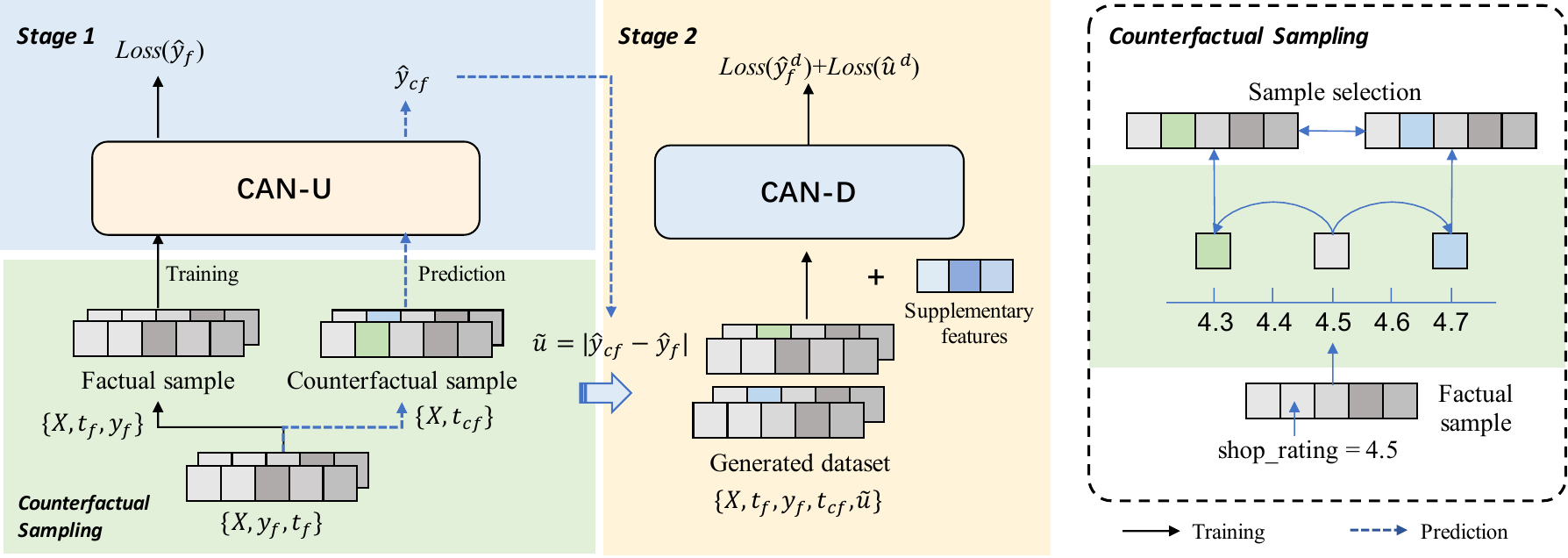}
  \caption{Diagram illustrating TSCAN's two-stage training process and counterfactual sampling diagram, where the black solid line represents the training flow and the blue dotted line represents the prediction flow.}
  \label{fig:TSCAN_training_process}
\end{figure}

The two-stage training process is proposed to solve the regularization-induced information loss demonstrated in Section~\ref{sec:representation_constraint}, where bias correction and final prediction are decoupled into two specialized stages, rather than compromising between these competing objectives within a single model (Figure~\ref{fig:TSCAN_training_process}). Stage 1 (CAN-U) focuses exclusively on generating unbiased pseudo-uplift labels through rigorous bias correction, while Stage 2 (CAN-D) leverages these labels to optimize predictive performance without regularization constraints.

\textbf{Stage 1:} 
In this stage, the CAN-U model is trained on the observed dataset $D = {\{(x_i, t_i, y_i)\}}_{i=1}^N$, incorporating the Representation Constraint Module to mitigate selection bias via dual regularization mechanisms~\cite{CFR_ISW2019}. The overall training objective is formulated as a minimax optimization problem that jointly accounts for factual outcome prediction, distributional balance, and adversarial propensity score regularization:
\begin{equation}
  \mathcal{L}_{\theta}=\mathcal{L}(\hat{y}_i,y_i)+\alpha \mathcal{L}_{\text{IPM}}+\beta \mathcal{R}(h)
\end{equation}
\begin{equation}
\min_{\theta} \max_{\pi} \left( \mathcal{L}_{\theta} - \lambda \mathcal{L}_{\pi} \right),
\end{equation}
where $\mathcal{L}(\hat{y}_i,y_i)$ denotes the factual outcome prediction loss, $\mathcal{L}_{\text{IPM}}$ is the Integral Probability Metric loss enforcing distributional balance between treatment groups, $\mathcal{L}_{\pi}$ is the propensity score estimation loss, and $\mathcal{R}(h)$ represents $\ell_2$ regularization. $\lambda$, $\alpha$ and $\beta$ are hyperparameters that control the relative importance of the adversarial propensity task, the IPM loss, and the $\ell_2$ regularization, respectively.

The adversarial training procedure follows a bilevel minimax strategy~\cite{AdversariallyBalanced2023}, which jointly optimizes factual outcome prediction, enforces distributional balance between treatment groups via IPM, and adversarially suppresses treatment-predictive information in the latent representation through propensity score estimation. Consequently, the learned representation exhibits improved alignment across treatment groups in the latent space, enhancing the robustness of individual treatment effect estimation~\cite{5,AdversariallyBalanced2023}.

\textbf{Stage 2:}
Given a factual observation $(X,t_{f},y_f)$, we construct a corresponding counterfactual instance $(X,t_{cf})$ by selecting an alternative treatment assignment $t_{cf} \neq t_f$. The counterfactual treatment $t_{cf}$ is sampled either uniformly at random from the support of the treatment variable or according to a predefined probabilistic strategy (e.g., based on empirical treatment distribution or domain heuristics).
Then, the pseudo-uplift label is derived as $\tilde{u}=\hat{y}(X,t_{cf})-\hat{y}(X,t_f)$ for each observation, where $\hat{y}(X,t)$ denotes the outcome predicted by the CAN-U model trained in Stage 1. This difference-based formulation aligns with theoretical results in causal inference~\cite{2008imben,2018chernozhukov}, which demonstrate that estimating treatment effects is statistically more efficient than estimating potential outcomes directly, particularly in high-dimensional settings.
The resulting complete and unbiased dataset $\tilde{D}=(X,t_{f},y_f,t_{cf},\tilde{u})$ enables CAN-D to be trained via a dual-loss objective:

\begin{equation}
\mathcal{L}_d=\mathcal{L}(y_f,\hat{y}_{f}^d)+\gamma\mathcal{L}(\tilde{u},\hat{u}^d)
\end{equation}
where the first term reinforces factual outcome prediction accuracy, while the second term transfers causal knowledge from CAN-U. The weight parameter $\gamma$ balances these objectives.

This two-stage architecture provides three critical advantages over conventional approaches:
\begin{enumerate}
    \item \textbf{Bias-Variance Trade-off Optimization}: By separating bias correction from final prediction, we avoid the regularization-induced variance inflation documented in~\cite{26,27}. CAN-D focuses exclusively on minimizing prediction error without constraints.
    
    \item \textbf{Error Correction Mechanism}: The factual outcome reinforcement term $\mathcal{L}(y_f,\hat{y}_{f}^d)$ enables CAN-D to correct systematic errors in predictions from CAN-U. Empirical validation is provided in Section~\ref{sec:experiments}.
    
    \item \textbf{Contextual Adaptation}: Without regularization constraints, CAN-D's context-aware attention layers can fully exploit contextual information that might have been suppressed by IPM regularization in Stage 1. This addresses the contextual underutilization problem highlighted in recent causal literature~\cite{huang2024entire}.
\end{enumerate}

\section{EMPIRICAL EVALUATIONS}\label{sec:experiments}
In this section, we design and conduct a series of comprehensive experiments to address the following three research questions: RQ1: How does the proposed TSCAN model perform compared to the baselines? RQ2: What is the contribution of each component in the model? RQ3: How does the TSCAN model perform in real-world online scenarios?

\subsection{Experimental Setup}

\subsubsection{Datasets}
We conduct experiments on two real-world datasets sourced from Taobao Shangou (previously called Ele.me), one of China's largest online food ordering platforms: 
\begin{itemize}
\item \textbf{Eleshop-1M}: This dataset contains online data of 1 million catering merchants. The treatment variable is continuous, defined as the merchant's average shop rating. The outcome is the total order count over a fixed period. Merchant features include operational attributes, such as shop ratings, operating hours, number of dishes and number of reviews. Contextual features capture the external market environment, such as business district type, regional user attributes, and the local supply-demand status at the time of observation.  

\item \textbf{Shop Activities}: This dataset contains data from 700k merchants. The treatment variable is binary, indicating whether a merchant participated in a specific ``new customer coupons'' marketing activity. The outcome is the total order count. Merchant features include menu categories, store exposure, and prices of main dishes. Contextual features include business district type, user type, time-period, average discount in the business district, and supply-demand status. We provide a detailed schema and summary statistics for both the Eleshop-1M and Shop Activities datasets in Table~\ref{tab:dataset_schema}.
\end{itemize}

\begin{table}[!ht]
\centering
\caption{Schema and summary statistics for the Eleshop-1M and Shop Activities datasets.}\label{tab:dataset_schema}
\begin{tabular}{lcc}
\toprule
\textbf{Attribute} & \textbf{Eleshop-1M} & \textbf{Shop Activities} \\
\midrule
\textbf{Task Type} & Continuous Treatment & Binary Treatment \\
\textbf{Total Samples} & 1,000,000 & 700,000 \\
\textbf{Train / Test Split} & 800k / 200k & 500k / 200k \\
\midrule 
\textbf{Merchant Features} & 42 & 48 \\
\textbf{Context Features} & 19 & 23 \\
\textbf{Total Input Features} & 61 & 71 \\
\midrule
\textbf{Treatment Variable} & average customer rating & participation in a new customer activity\\
\textbf{Outcome Variable} & order count (continuous) & order count (continuous) \\
\bottomrule
\end{tabular}
\end{table}

\subsubsection{Evaluation Metrics}
We evaluate the performance of TSCAN and other baseline models using two widely-recognized metrics (QINI and AUUC) and two contextualized derived metrics (CAUUC and CQINI). In addition, we include a key visualization tool (Gain Curve) to enable a more intuitive comparison of model performance.

\begin{itemize}
\item \textbf{Normalized QINI and AUUC}: QINI evaluates the effectiveness of uplift models in distinguishing between subsets of a population that respond differently to a treatment~\cite{2025qini}. AUUC provides a standardized performance measure that reflects the model's ability to accurately identify and segment the population based on their uplift potential~\cite{Liu2024BenchmarkingFD}. 

\item \textbf{Context-wise AUUC (CAUUC) and Context-wise QINI (CQINI)}: Standard AUUC and QINI provide global performance summaries but may obscure significant variations in model behavior across different operational contexts (e.g., high vs low supply–demand environments). When evaluating AUUC, uplift values from different contexts are mixed together, making it impossible to evaluate the ranking effect of the same context accurately. Therefore, to rigorously evaluate a model’s ability to capture context-dependent heterogeneity in treatment effects, this paper proposes two context-stratified variants: CAUUC and CQINI. These metrics compute a weighted average of AUUC/QINI scores across merchant subgroups defined by distinct contextual conditions, thereby offering a more granular assessment of contextual sensitivity and robustness. 

Specifically, CAUUC is the weighted average of AUUC for merchant groups across different contexts:
\begin{equation}
  \mathrm{CAUUC} = \frac{\sum_{g=1}^{G} N_g \cdot \mathrm{AUUC}_g}{\sum_{g=1}^{G} N_g}
\end{equation}
where $G$ is the number of merchant groups stratified by different contexts, $N_g$ is the sample count of group $g$, and $\text{AUUC}_g$ is the AUUC value of group $g$.

CQINI is the weighted average of QINI for merchant groups across different contexts:
\begin{equation}
  \mathrm{CQINI}=\frac{\sum_{g=1}^{G}{N_g \cdot \mathrm{QINI}_g}}{\sum_{g=1}^{G}{N_g}}
\end{equation}
where $\text{QINI}_g$ is the QINI value of group $g$. In these experiments, contextual groups are defined by business district type and time-period. For continuous treatments, we discretize them into multiple intervals and compute the average AUUC and average QINI for each interval. 

\item \textbf{Gain Curve}: The Gain Curve is a visualization tool used to evaluate the effectiveness of uplift models. It plots the cumulative uplift against the proportion of the population targeted, ranked by predicted uplift scores. A steeper Gain Curve indicates superior model performance, as it shows that the model can achieve higher uplift by targeting a smaller subset of the population.

\end{itemize}

\subsubsection{Baselines and Parameter Settings}
To comprehensively evaluate the proposed TSCAN framework, we select representative methods spanning three major categories of uplift modeling approaches: meta-learners, Tree-based methods and Deep learning methods.
\begin{itemize} 
    \item \textbf{S-Learner}~\cite{10}: A meta-learner that estimates treatment effects by including treatment as a feature in a single model.
    \item \textbf{T-Learner}~\cite{10}: A meta-learner that trains separate models for treatment and control groups and computes their difference. 
    \item \textbf{X-Learner}~\cite{10}: An extension of T-Learner that incorporates propensity scores and cross-fitting to reduce bias in imbalanced settings. 
    \item \textbf{BART}~\cite{11}: It employs a sum-of-trees approach with regularization to estimate heterogeneous treatment effects.
    \item  \textbf{Causal Forest}~\cite{12}: It extends random forests with specific splitting criteria designed for causal inference.
    \item \textbf{TarNet}~\cite{20}: It learns balanced, treatment-invariant representations of covariates by using two separate prediction heads for treatment and control conditions, while sharing a common representation layer; 
    \item  \textbf{DragonNet}~\cite{9}: This model jointly learns a shared representation of covariates and estimates potential outcomes under treatment and control conditions using a modified architecture inspired by the TarNet framework; 
    \item \textbf{TransTEE}~\cite{5}: This approach introduces transformer architecture for uplift modeling with explicit treatment representation; 
    \item  \textbf{EFIN}~\cite{14}: It designs a treatment-gated feature interaction network to capture heterogeneous treatment effects; 
    \item  \textbf{CFR-ISW}~\cite{CFR_ISW2019}: It combines representation balancing with importance sampling weighting; 
    \item  \textbf{DESCN}~\cite{2022DESCN}: This model captures the integrated information of treatment and response through a cross network in a multi-task learning manner, and it employs an intermediate pseudo treatment effect prediction network to relieve sample imbalance.
\end{itemize}

In Stage 1, CAN-U is trained with early stopping using Adam optimizer (learning rate=0.015, $\beta_{1}$=0.9, $\beta_{2}$=0.999). The regularization weights are set as $\lambda$=0.5 (adversarial task), $\alpha$ =0.01 (IPM loss), and $\beta$ =1e-5 ($\ell_2$ regularization). Subsequently, for each sample in the dataset, we generate its corresponding counterfactual counterpart. Then the trained CAN-U model is applied to these original–counterfactual outcome pairs to estimate the uplift label for each instance. In Stage 2, CAN-D is trained with early stopping using the same optimizer settings on the dataset constructed from Stage 1. The uplift prediction loss weight $\gamma$ is set to 0.6 through grid search on the validation set. 

Since DragonNet, TarNet, and DESCN do not natively support continuous treatments, we extend their architectures to handle continuous treatments on the Eleshop-1M dataset by replacing the dual-head output layer with $K=5$ parallel heads, following the approach of~\cite{14,5}. Since X-Learner is difficult to extend, we do not report its results on this dataset.  
All models are implemented using Python 3.8 and PyTorch. We employ the Maximum Mean Discrepancy (MMD) as the IPM loss. All experiments are repeated five times, and the results are averaged.

\subsection{Overall Performance Assessment}

\subsubsection{RQ1: How does the proposed TSCAN model perform compared to the baseline models?}
The performance of TSCAN compared with the baseline models on the datasets Eleshop-1M and Shop Activities is shown in Table~\ref{tab:t1}. From the results, we can draw the following conclusions:

(1) TSCAN achieves the best performance on both the Eleshop-1M and Shop Activities datasets. On the Eleshop-1M dataset with continuous treatments, TSCAN surpasses the best baseline (TransTEE) by 0.0049 in AUUC and 0.0153 in CAUUC. On the Shop Activities dataset with binary treatments, TSCAN outperforms DESCN by +0.0033 in AUUC and +0.0080 in QINI. This consistent superiority across different treatment types validates TSCAN's flexibility and robustness. The performance gains on context-aware metrics (CAUUC/CQINI) confirm that the Context-Aware Attention Layer effectively captures context-dependent treatment effects. This advantage is amplified on the Eleshop-1M dataset with continuous treatments, highlighting the benefit of the isotonic output layer for modeling incremental treatment effects.

(2) Deep learning models generally outperform traditional meta-learners and tree-based approaches, validating the crucial role of complex feature interactions in merchant diagnosis scenarios. Notably, within the deep learning category, models explicitly designed for treatment effect estimation (TransTEE, CFR-ISW and EFIN) achieve higher performance than S-Learner and T-Learner, confirming that specialized causal architectures yield more accurate uplift estimates. However, TSCAN's two-stage training strategy provides a significant advantage over all these approaches, as it avoids the information loss caused by regularization-induced bias while maintaining causal robustness.

(3) Significant performance differences exist between models when handling different treatment types. For continuous treatments (Eleshop-1M dataset), TransTEE demonstrates stronger performance among baselines (AUUC=0.7489), leveraging its transformer architecture to model dose-response relationships effectively. For binary treatments (Shop Activities), DESCN achieves competitive results (QINI=0.0974). DragonNet performs consistently well across both treatment types (AUUC=0.7202 for continuous and 0.6301 for binary), demonstrating the effectiveness of its adversarial balancing approach. By contrast, the performance of tree-based models varies significantly depending on the treatment type, suggesting they are particularly well-suited to binary treatments. TSCAN, however, maintains strong performance across both treatment types, demonstrating the versatility of the proposed architecture.

\begin{figure}[h]
  \centering
  \begin{subfigure}[b]{0.46\textwidth}
    \centering
    \includegraphics[width=\textwidth]{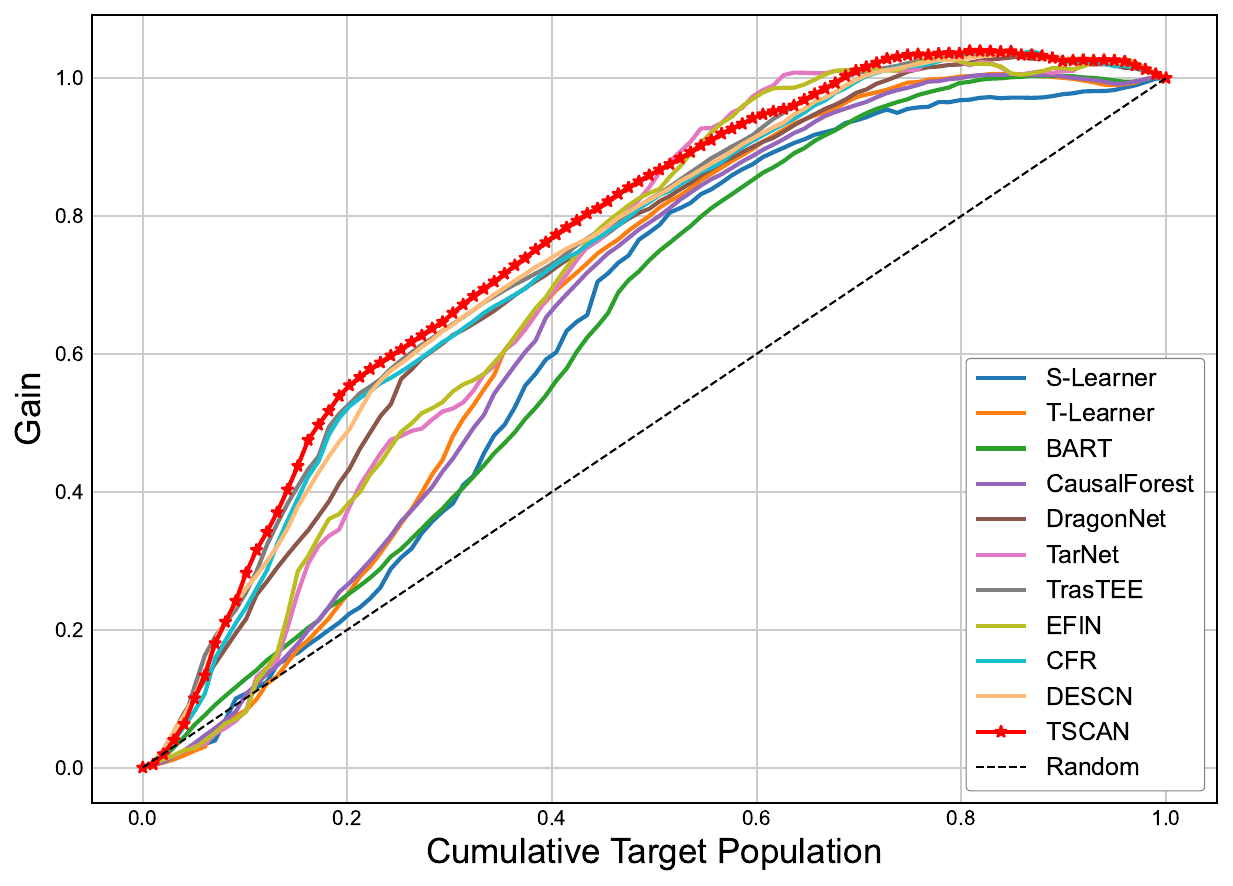}
    \caption{Gain curve on the Eleshop-1M.}
    \label{fig:evalResult_eleme1m}
  \end{subfigure}
  \hspace{1em} 
  \begin{subfigure}[b]{0.46\textwidth}
    \centering
    \includegraphics[width=\textwidth]{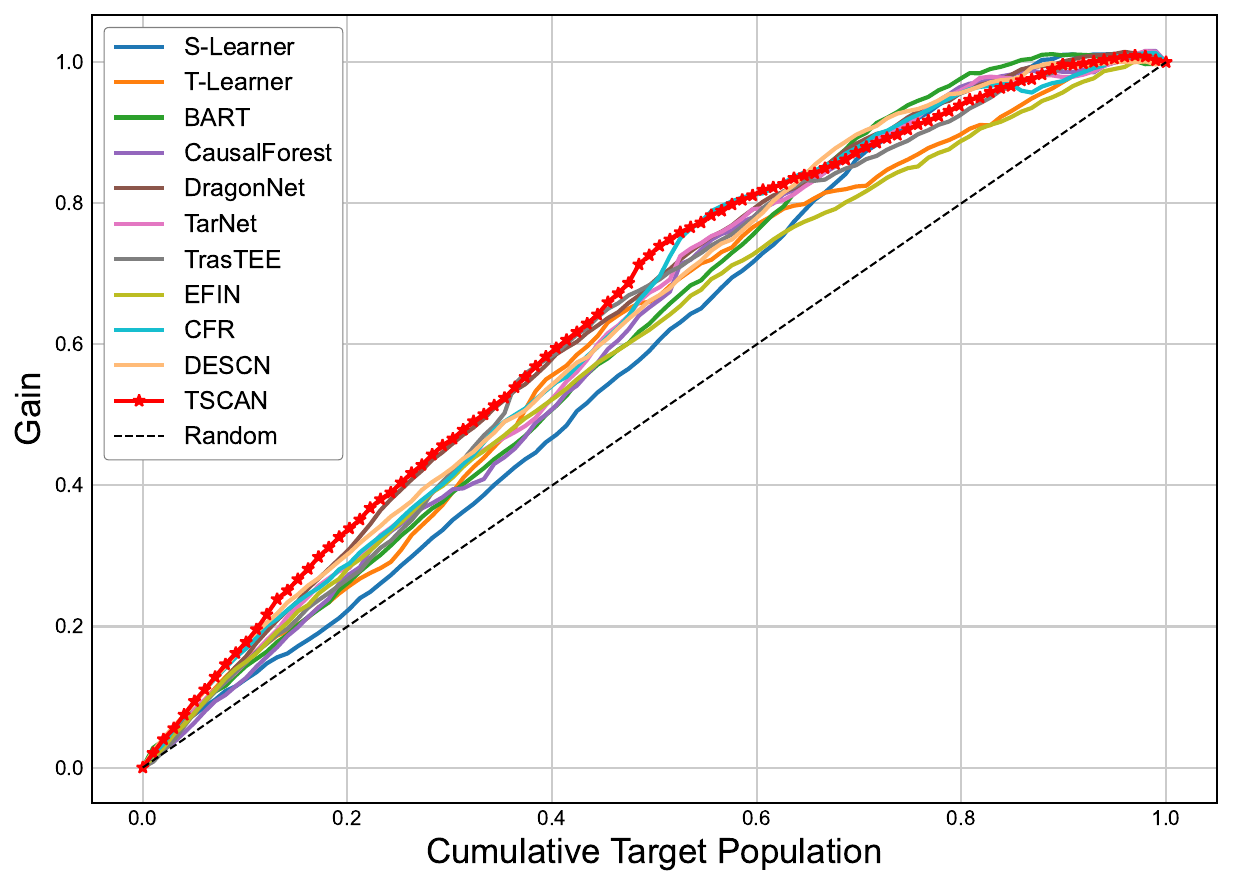}
    \caption{Gain curve on the Shop Activities.}
    \label{fig:evalResult_shopactivity}
  \end{subfigure}
  \caption{Gain curves for uplift prediction on benchmark datasets.}
  \label{fig:evalResult}
\end{figure}

To intuitively compare the cumulative uplift effects across different models, the Gain Curves of various approaches on the benchmarks are demonstrated in Figure~\ref{fig:evalResult}. In Gain Curve, a steeper trajectory reaching higher uplift faster indicates better model performance. The ideal curve approaches the upper-left corner, while the diagonal (black dotted curve) represents random ordering. As shown in Figure~\ref{fig:evalResult}, TSCAN (red curve) exhibits the steepest ascent among the compared models, confirming its enhanced capability to identify high-uplift merchants through effective causal effect prioritization.

\begin{table}[!ht]
  \centering
  \caption{Model performance comparison on the two datasets}\label{tab:t1}
  \begin{tabular}{l|rrrr|rrrr}
    \toprule
    Dataset & \multicolumn{4}{c|}{Eleshop-1M} & \multicolumn{4}{c}{Shop Activities} \\ 
    \hline
    Metrics & CQINI & QINI & CAUUC & AUUC & CQINI & QINI & CAUUC & AUUC  \\ 
    \midrule
    S-Learner & 0.1708 & 0.1836 & 0.6442 & 0.6649 & 0.0874 & 0.0828 & 0.5819 & 0.5800  \\ 
    T-Learner & 0.2026 & 0.2163 & 0.6751 & 0.6871 & 0.0881 & 0.0842 & 0.6106 & 0.5955  \\
    X-Learner & \multicolumn{1}{c}{--} & \multicolumn{1}{c}{--} & \multicolumn{1}{c}{--} & \multicolumn{1}{c}{--} & 0.0890 & 0.0886 & 0.6164 & 0.6068 \\
    \hline
    BART & 0.1236 & 0.1615 & 0.6242 & 0.6420 & 0.0914 & 0.0861 & 0.6196 & 0.6096  \\ 
    Causal Forest & 0.1887 & 0.2014 & 0.6630 & 0.6785 & 0.0889 & 0.0873 & 0.6142 & 0.6030  \\
    \hline
    TarNet & 0.2323 & 0.2248 & 0.7142 & 0.7079 & 0.0936 & 0.0969 & \underline{0.6335} & 0.6283  \\
    DragonNet & 0.2470 & 0.2393 & 0.7266 & 0.7202 & 0.0930 & 0.0953	& 0.6329 & \underline{0.6301}  \\
    TransTEE & \underline{0.2652} & \underline{0.2608} & \underline{0.7533} & \underline{0.7489} & 0.0919 & 0.0935 & 0.6270 & 0.6256  \\
    EFIN & 0.2344 & 0.2267 & 0.7162 & 0.7107 & 0.0828 & 0.0837 & 0.5947 & 0.5949  \\
    CFR-ISW & 0.2536 & 0.2489 & 0.7434 & 0.7388 & 0.0906 & 0.0925 & 0.6217 & 0.6167  \\
    DESCN & 0.2581 & 0.2524 & 0.7489 & 0.7442 & \underline{0.0938} & \underline{0.0974} & 0.6325 & 0.6295  \\
    \hline
    TSCAN (ours) & \textbf{0.2839} & \textbf{0.2687} & \textbf{0.7686} & \textbf{0.7538} & \textbf{0.0994} & \textbf{0.1054} & \textbf{0.6379} & \textbf{0.6328} \\
    \bottomrule
  \end{tabular}
\end{table}

\subsubsection{RQ2: What is the contribution of each component in the model?}
We conduct ablation studies to quantify the contribution of four core components in TSCAN: 
\begin{itemize} 
    \item \textbf{Two-stage training strategy}: To isolate the effect of decoupled bias correction, we compare the full TSCAN model (CAN-D submodel trained in Stage 2) against CAN-U (the Stage 1 output).

    \item \textbf{Context-Aware Attention layer}: To evaluate whether explicit context-merchant interaction modeling outperforms naive feature fusion, we create TSCAN-RA by replacing the Context-Aware Attention layer in both submodels with a standard fully-connected layer that treats context features as ordinary merchant features.

    \item \textbf{Contextual features}: To measure the necessity of contextual information for treatment effect estimation, we construct TSCAN-RC by removing all contextual features from the input.

    \item \textbf{Isotonic output layer}: To assess the impact of isotonic output layer on uplift estimation, we develop TSCAN-RISO by replacing the isotonic output layer in CAN-D with a standard fully-connected layer, thereby reducing the dual-loss objective to outcome prediction loss.
\end{itemize}

The experimental results are presented in Table~\ref{tab:t2}. Based on the experimental results, the following conclusions can be established: 

(1) The two-stage training strategy improves performance: CAN-D (full TSCAN) outperforms CAN-U on both datasets, with relative improvements of up to 5.95\% in QINI and 1.45\% in AUUC on the Eleshop-1M dataset. This validates the hypothesis that decoupling bias correction from prediction avoids the information loss inherent in single-stage regularization approaches. The performance gap between CAN-U and CAN-D is larger on CAUUC and CQINI metrics (CAUUC difference of 0.0212 vs. AUUC difference of 0.0108), indicating that removing regularization constraints particularly benefits context-sensitive estimation.

(2) Contextual information is crucial: removing contextual features (TSCAN-RC) causes performance degradation (AUUC drops by 0.0856 on Eleshop-1M), confirming the hypothesis that treatment effects are highly context-dependent. Second, the Context-Aware Attention Layer provides significant gains over treating context as ordinary features (TSCAN-RA vs. TSCAN), particularly on context-stratified metrics (CAUUC improves by 0.0306 on Eleshop-1M), demonstrating its effectiveness in modeling merchant-treatment-context interactions.

(3) The isotonic output layer is critical for accurate uplift estimation. TSCAN-RISO shows performance degradation compared to full TSCAN (AUUC drops by 0.0162), particularly on continuous treatment tasks. By modeling the uplift value directly, CAN-D is able to focus on the supervised learning of uplift labels while maintaining the accuracy of factual outcome prediction.

\begin{table}[!ht]
  \centering
  \caption{Model performance comparison of TSCAN-RC, TSCAN-RA, TSCAN-RISO, CAN-U and TSCAN (the CAN-D sub-model)}\label{tab:t2}
  \begin{tabular}{l|rrrr|rrrr}
    \toprule
    Dataset & \multicolumn{4}{c|}{Eleshop-1M} & \multicolumn{4}{c}{Shop Activities} \\ 
      \hline
      Metrics & CQINI & QINI & CAUUC & AUUC & CQINI & QINI & CAUUC & AUUC  \\
      \midrule
      TSCAN-RC & 0.1737 & 0.1847 & 0.6536 & 0.6682 & 0.0865 & 0.0812 & 0.5812 & 0.5735  \\
      TSCAN-RA & 0.2526 & 0.2429 & 0.7380 & 0.7334 & 0.0939 & 0.0972 & 0.6234 & 0.6302  \\
      TSCAN-RISO & 0.2602 & 0.2490 & 0.7452 & 0.7376 & 0.0933 & 0.0969 & 0.6227 & 0.6296   \\
      CAN-U & 0.2639 & 0.2536 & 0.7474 & 0.7430 & 0.0941 & 0.0977 & 0.6283 & 0.6309  \\
      \hline
      TSCAN (CAN-D) & \textbf{0.2839} & \textbf{0.2687} & \textbf{0.7686} & \textbf{0.7538} & \textbf{0.0994} & \textbf{0.1054} & \textbf{0.6379} & \textbf{0.6328} \\
      \bottomrule
  \end{tabular}
\end{table}

\begin{figure}[h]
  \centering
  \includegraphics[width=0.48\textwidth]{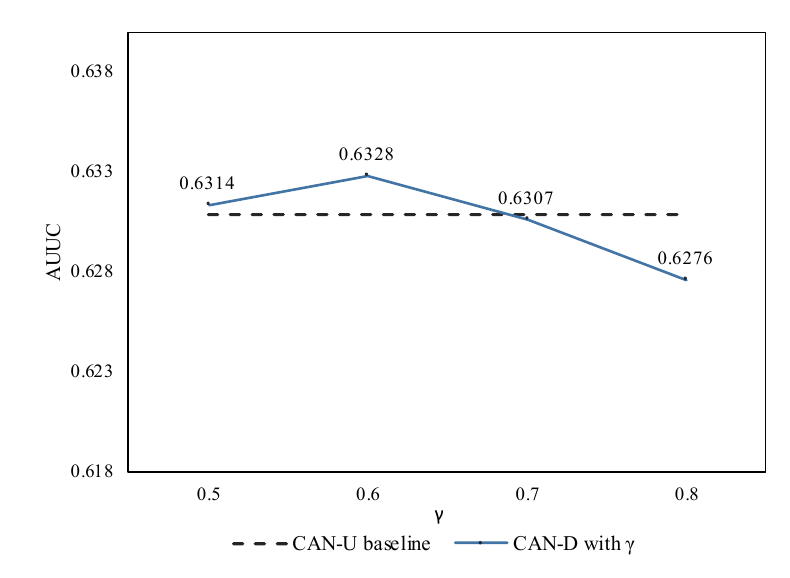}
  \includegraphics[width=0.48\textwidth]{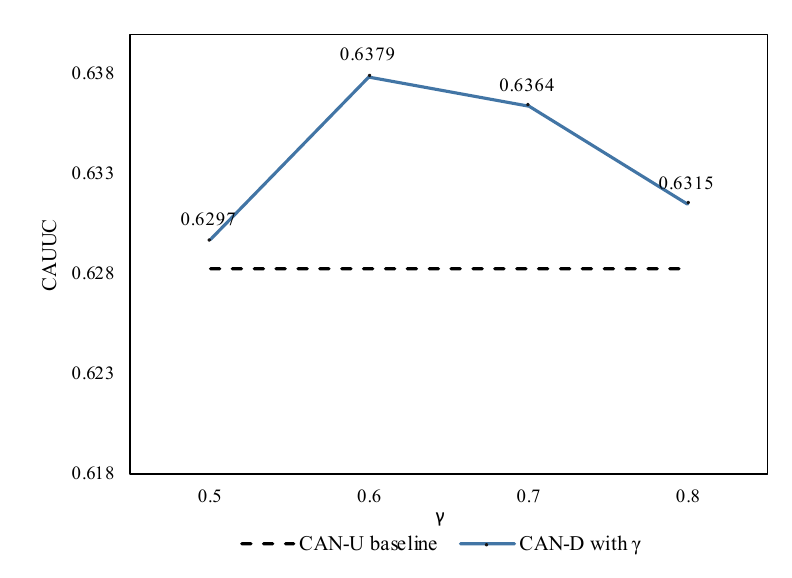}
  \caption{Performance of CAN-D under different values of the uplift prediction loss weight $\gamma$}\label{fig:gamma_auuc_cauuc}
\end{figure}

To investigate the influence of the uplift prediction loss weight $\gamma$ on the CAN-D model, we conduct a series of comparative experiments using different values of $\gamma$, with AUUC and CAUUC serving as the primary evaluation metrics. The results are presented in Figure~\ref{fig:gamma_auuc_cauuc}. As $\gamma$ varies from 0.5 to 0.8, both AUUC and CAUUC initially increase and subsequently decline, indicating the sensitivity of model performance to this hyperparameter. The optimal value is found to be approximately $\gamma$=0.6, yielding an AUUC of 0.6328 and a CAUUC of 0.6379. For reference, the black dashed line denotes the performance of the CAN-U baseline (AUUC: 0.6309; CAUUC: 0.6283). At $\gamma$=0.6, CAN-D achieves a relative improvement of 0.0019 (0.3\%) in AUUC and 0.0096 (1.5\%) in CAUUC over CAN-U. These results demonstrate that the proposed two-stage training strategy effectively enhances the model’s counterfactual prediction accuracy and its capacity to leverage contextual information.

\subsubsection{RQ3: How does the TSCAN model perform in real-world online scenarios?}
To evaluate the performance of TSCAN in real-world online scenarios, we deployed TSCAN on a real merchant diagnosis system of an online food ordering platform in China. In this application, we estimate the increase in order volume (the outcome) after merchants adopt specific suggestions, such as configuring discount vouchers for new customers and setting up shop posters. The A/B test compares TSCAN against BART (the previously deployed model) across 90,000 merchants randomly assigned to treatment groups. Each merchant received personalized diagnostic suggestions ranked by predicted uplift. As shown in Table~\ref{tab:t3}, in the online experiment, TSCAN outperforms the baseline model BART, achieving an AUUC improvement of 0.0349, a CAUUC improvement of 0.0411 and a 0.76\% increase in order volume (95\% CI [0.68\%, 0.84\%], \textit{p}=0.001). Manual analysis of high-performing market segments reveals that TSCAN effectively identifies context-specific opportunities: during peak demand periods in business districts, it correctly down-weights discount suggestions and up-weights exposure-boosting suggestions; conversely, during low demand periods in residential areas, it recommends targeted discounts. This adaptive behavior validates the design of the context-aware attention layer.

\begin{table}[!ht]
  \centering
  \caption{Online performance comparison of TSCAN and baseline model BART}\label{tab:t3}
  \begin{tabular}{lrrr}
  \toprule
      Metrics & CAUUC & AUUC & Order increase  \\ 
      \midrule
      Base (BART) & 0.6331 & 0.6370 & 0.00\%  \\
      TSCAN & 0.6742 & 0.6719 & +0.76\% \\
      \bottomrule
  \end{tabular}
\end{table}

\section{Conclusion}
In this paper, we introduce TSCAN, a Context-Aware uplift model based on a two-stage training approach. TSCAN effectively mitigates the negative impacts of traditional regularization methods such as IPM loss and propensity score prediction by employing a two-stage training strategy with CAN-U and CAN-D models. Additionally, by integrating a Context-Aware attention layer, TSCAN leverages contextual features to enhance the accuracy of treatment effect estimation across diverse environments. Through extensive experiments on two large-scale real-world datasets and a live deployment on one of China's largest food ordering platforms, we demonstrate that TSCAN achieves consistent improvements across multiple evaluation metrics. Despite its empirical success, our approach has several limitations. First, the two-stage training process increases computational complexity and training time compared to single-stage models. Second, while our experiments span both continuous and binary treatments, the datasets originate from the online food ordering industry, which may limit the generalizability of the proposed findings to substantially different domain contexts or treatment types. Finally, the isotonic output layer assumes monotonic treatment effects within discretized intervals, which may not hold for treatments with non-monotonic dose-response relationships.

There are several potential future works for exploration. We will investigate distillation techniques to compress the two-stage architecture into an end-to-end model without significant performance degradation. Second, we plan to develop unsupervised clustering techniques to automatically identify latent contextual segments from time-series merchant behavior data. Finally, we will validate TSCAN across diverse e-commerce domains to assess its transferability and robustness. These improvements will further bridge the gap between causal machine learning theory and practical business applications.


\bibliographystyle{ACM-Reference-Format}
\bibliography{reference}

@misc{1,
  author       = {David Curry},
  title        = {Food Delivery App Revenue and Usage Statistics (2025)},
  howpublished = {Website},
  year         = {2025},
  note         = {https://www.businessofapps.com/data/food-delivery-app-market/.}
}

@article{2,
  title    = {Does online food delivery improve the equity of food accessibility? A case study of Nanjing, China},
  journal  = {Journal of Transport Geography},
  volume   = {107},
  pages    = {103516},
  year     = {2023},
  issn     = {0966-6923},
  doi      = {https://doi.org/10.1016/j.jtrangeo.2022.103516},
  url      = {https://www.sciencedirect.com/science/article/pii/S0966692322002393},
  author   = {Shanqi Zhang and Hui Luan and Feng Zhen and Yu Kong and Guangliang Xi}
}

@article{3,
  title    = {Balance between profit and fairness: Regulation of online food delivery OFD platforms},
  journal  = {International Journal of Production Economics},
  volume   = {269},
  pages    = {109144},
  year     = {2024},
  issn     = {0925-5273},
  doi      = {https://doi.org/10.1016/j.ijpe.2024.109144},
  url      = {https://www.sciencedirect.com/science/article/pii/S092552732400001X},
  author   = {Xu Ji and Xuerong Li and Shouyang Wang}
}

@misc{4,
  title         = {A unified survey of treatment effect heterogeneity modeling and uplift modeling},
  author        = {Weijia Zhang and Jiuyong Li and Lin Liu},
  year          = {2021},
  eprint        = {2007.12769},
  archiveprefix = {arXiv},
  primaryclass  = {stat.ME},
  url           = {https://arxiv.org/abs/2007.12769}
}

@article{5,
  title   = {Exploring Transformer Backbones for Heterogeneous Treatment Effect Estimation},
  author  = {Yi-Fan Zhang and Hanlin Zhang and Zachary Chase Lipton and Li Erran Li and Eric P. Xing},
  journal = {Trans. Mach. Learn. Res.},
  year    = {2022},
  volume  = {2023},
  url     = {https://api.semanticscholar.org/CorpusID:249151761}
}

@article{6,
  author     = {Yao, Liuyi and Chu, Zhixuan and Li, Sheng and Li, Yaliang and Gao, Jing and Zhang, Aidong},
  title      = {A Survey on Causal Inference},
  year       = {2021},
  issue_date = {October 2021},
  publisher  = {Association for Computing Machinery},
  address    = {New York, NY, USA},
  volume     = {15},
  number     = {5},
  issn       = {1556-4681},
  url        = {https://doi.org/10.1145/3444944},
  doi        = {10.1145/3444944},
  journal    = {ACM Trans. Knowl. Discov. Data},
  month      = may,
  articleno  = {74},
  numpages   = {46}
}

@article{7,
  title   = {A survey of deep causal models and their industrial applications},
  author  = {Zongyu Li and Zheng Hua Zhu and Xiaoning Guo and Shuai Zheng and Zhenyu Guo and Siwei Qiang and Yao Zhao},
  journal = {Artif. Intell. Rev.},
  year    = {2022},
  volume  = {57},
  pages   = {298},
  url     = {https://api.semanticscholar.org/CorpusID:253523500}
}

@inproceedings{8,
  title     = {Causal Effect Inference with Deep Latent-Variable Models},
  author    = {Christos Louizos and Uri Shalit and Joris M. Mooij and David A. Sontag and Richard S. Zemel and Max Welling},
  booktitle = {Neural Information Processing Systems},
  year      = {2017},
  url       = {https://api.semanticscholar.org/CorpusID:260564}
}

@inbook{9,
  author    = {Shi, Claudia and Blei, David M. and Veitch, Victor},
  title     = {Adapting neural networks for the estimation of treatment effects},
  year      = {2019},
  publisher = {Curran Associates Inc.},
  address   = {Red Hook, NY, USA},
  booktitle = {Proceedings of the 33rd International Conference on Neural Information Processing Systems},
  articleno = {225},
  numpages  = {11}
}

@article{10,
  title     = {Metalearners for estimating heterogeneous treatment effects using machine learning},
  volume    = {116},
  issn      = {1091-6490},
  url       = {http://dx.doi.org/10.1073/pnas.1804597116},
  doi       = {10.1073/pnas.1804597116},
  number    = {10},
  journal   = {Proceedings of the National Academy of Sciences},
  publisher = {Proceedings of the National Academy of Sciences},
  author    = {Künzel, Sören R. and Sekhon, Jasjeet S. and Bickel, Peter J. and Yu, Bin},
  year      = {2019},
  month     = feb,
  pages     = {4156–4165}
}

@article{11,
  title     = {BART: Bayesian additive regression trees},
  volume    = {4},
  issn      = {1932-6157},
  url       = {http://dx.doi.org/10.1214/09-AOAS285},
  doi       = {10.1214/09-aoas285},
  number    = {1},
  journal   = {The Annals of Applied Statistics},
  publisher = {Institute of Mathematical Statistics},
  author    = {Chipman, Hugh A. and George, Edward I. and McCulloch, Robert E.},
  year      = {2010},
  month     = mar
}

@article{12,
  author  = {Ramachandra, Vikas and Athey, Susan and Wager, Stefan},
  year    = {2015},
  month   = {10},
  pages   = {},
  title   = {Estimation and Inference of Heterogeneous Treatment Effects using Random Forests},
  volume  = {113},
  journal = {Journal of the American Statistical Association},
  doi     = {10.1080/01621459.2017.1319839}
}

@inproceedings{CFR_ISW2019,
  title     = {CounterFactual Regression with Importance Sampling Weights},
  author    = {Hassanpour, Negar and Greiner, Russell},
  booktitle = {Proceedings of the Twenty-Eighth International Joint Conference on
               Artificial Intelligence, IJCAI-19},
  publisher = {International Joint Conferences on Artificial Intelligence Organization},
  pages     = {5880--5887},
  year      = {2019},
  month     = {7},
  doi       = {10.24963/ijcai.2019/815},
  url       = {https://doi.org/10.24963/ijcai.2019/815}
}

@inproceedings{14,
  author    = {Liu, Dugang and Tang, Xing and Gao, Han and Lyu, Fuyuan and He, Xiuqiang},
  title     = {Explicit Feature Interaction-aware Uplift Network for Online Marketing},
  year      = {2023},
  isbn      = {9798400701030},
  publisher = {Association for Computing Machinery},
  address   = {New York, NY, USA},
  url       = {https://doi.org/10.1145/3580305.3599820},
  doi       = {10.1145/3580305.3599820},
  booktitle = {Proceedings of the 29th ACM SIGKDD Conference on Knowledge Discovery and Data Mining},
  pages     = {4507–4515},
  numpages  = {9},
  location  = {Long Beach, CA, USA},
  series    = {KDD '23}
}

@inproceedings{15,
  author    = {Liu, Ruoqi and Yin, Changchang and Zhang, Ping},
  booktitle = {2020 IEEE International Conference on Data Mining ICDM},
  title     = {Estimating Individual Treatment Effects with Time-Varying Confounders},
  year      = {2020},
  volume    = {},
  number    = {},
  pages     = {382-391},
  doi       = {10.1109/ICDM50108.2020.00047},
  issn      = {2374-8486},
  month     = {Nov}
}

@inproceedings{16,
  author    = {Johansson, Fredrik D. and Shalit, Uri and Sontag, David},
  title     = {Learning representations for counterfactual inference},
  year      = {2016},
  publisher = {JMLR.org},
  booktitle = {Proceedings of the 33rd International Conference on International Conference on Machine Learning - Volume 48},
  pages     = {3020–3029},
  numpages  = {10},
  location  = {New York, NY, USA},
  series    = {ICML'16},
  volume    = {48},
  address   = {New York, New York, USA},
  month     = {20--22 Jun}
}

@article{17,
  title   = {Estimation of Individual Treatment Effect in Latent Confounder Models via Adversarial Learning},
  author  = {Changhee Lee and Nicholas Mastronarde and Mihaela van der Schaar},
  journal = {ArXiv},
  year    = {2018},
  volume  = {abs/1811.08943},
  url     = {https://api.semanticscholar.org/CorpusID:53716914}
}

@inproceedings{18,
  author    = {Jinsung Yoon and
               James Jordon and
               Mihaela van der Schaar},
  title     = {GANITE: Estimation of Individualized Treatment Effects using Generative
               Adversarial Nets},
  booktitle = {6th International Conference on Learning Representations, ICLR 2018,
               Vancouver, BC, Canada, April 30 - May 3, 2018, Conference Track Proceedings},
  publisher = {OpenReview.net},
  year      = {2018},
  url       = {https://openreview.net/forum?id=ByKWUeWA-},
  bibsource = {dblp computer science bibliography, https://dblp.org}
}

@inproceedings{19,
  author    = {Bica, Ioana and Jordon, James and van der Schaar, Mihaela},
  title     = {Estimating the effects of continuous-valued interventions using generative adversarial networks},
  year      = {2020},
  isbn      = {9781713829546},
  publisher = {Curran Associates Inc.},
  address   = {Red Hook, NY, USA},
  booktitle = {Proceedings of the 34th International Conference on Neural Information Processing Systems},
  articleno = {1379},
  numpages  = {12},
  location  = {Vancouver, BC, Canada},
  series    = {NIPS '20}
}

@inproceedings{20,
  author    = {Shalit, Uri and Johansson, Fredrik D. and Sontag, David},
  title     = {Estimating individual treatment effect: generalization bounds and algorithms},
  year      = {2017},
  publisher = {JMLR.org},
  booktitle = {Proceedings of the 34th International Conference on Machine Learning - Volume 70},
  pages     = {3076–3085},
  numpages  = {10},
  location  = {Sydney, NSW, Australia},
  series    = {ICML'17}
}

@misc{21,
  title         = {Covariate balancing using the integral probability metric for causal inference},
  author        = {Insung Kong and Yuha Park and Joonhyuk Jung and Kwonsang Lee and Yongdai Kim},
  year          = {2023},
  eprint        = {2305.13715},
  archiveprefix = {arXiv},
  primaryclass  = {stat.ML},
  url           = {https://arxiv.org/abs/2305.13715},
  booktitle     = {In International Conference on Machine Learning},
  pages         = {17430–17461},
  series        = {PMLR, 2023}
}

@inproceedings{23,
  author    = {Hatt, Tobias and Feuerriegel, Stefan},
  title     = {Estimating Average Treatment Effects via Orthogonal Regularization},
  year      = {2021},
  isbn      = {9781450384469},
  publisher = {Association for Computing Machinery},
  address   = {New York, NY, USA},
  url       = {https://doi.org/10.1145/3459637.3482339},
  doi       = {10.1145/3459637.3482339},
  booktitle = {Proceedings of the 30th ACM International Conference on Information \& Knowledge Management},
  pages     = {680–689},
  numpages  = {10},
  location  = {Virtual Event, Queensland, Australia},
  series    = {CIKM '21}
}

@article{24,
  title   = {Learning Decomposed Representation for Counterfactual Inference},
  author  = {Anpeng Wu and Kun Kuang and Junkun Yuan and Bo Li and Pan Zhou and Jianrong Tao and Qiang Zhu and Yueting Zhuang and Fei Wu},
  journal = {ArXiv},
  year    = {2020},
  volume  = {abs/2006.07040},
  url     = {https://api.semanticscholar.org/CorpusID:219636246}
}

@inproceedings{25,
  title     = {Multi-Cause Effect Estimation with Disentangled Confounder Representation},
  author    = {Ma, Jing and Guo, Ruocheng and Zhang, Aidong and Li, Jundong},
  booktitle = {Proceedings of the Thirtieth International Joint Conference on
               Artificial Intelligence, IJCAI-21},
  publisher = {International Joint Conferences on Artificial Intelligence Organization},
  editor    = {Zhi-Hua Zhou},
  pages     = {2790--2796},
  year      = {2021},
  month     = {8},
  note      = {Main Track},
  doi       = {10.24963/ijcai.2021/384},
  url       = {https://doi.org/10.24963/ijcai.2021/384}
}

@article{26,
  title   = {Counterfactual Representation Learning with Balancing Weights},
  author  = {Serge Assaad and Shuxi Zeng and Chenyang Tao and Shounak Datta and Nikhil Mehta and Ricardo Henao and Fan Li and Lawrence Carin},
  journal = {ArXiv},
  year    = {2020},
  volume  = {abs/2010.12618},
  url     = {https://api.semanticscholar.org/CorpusID:225067078}
}

@inproceedings{27,
  title     = {Limits of Estimating Heterogeneous Treatment Effects: Guidelines for Practical Algorithm Design},
  author    = {Alaa, Ahmed and van der Schaar, Mihaela},
  booktitle = {Proceedings of the 35th International Conference on Machine Learning},
  pages     = {129-138},
  year      = {2018},
  editor    = {Dy, Jennifer and Krause, Andreas},
  volume    = {80},
  series    = {Proceedings of Machine Learning Research},
  month     = {10--15 Jul},
  publisher = {PMLR},
  url       = {https://proceedings.mlr.press/v80/alaa18a.html}
}

@article{28,
  author  = {Johansson, Fredrik and Kallus, Nathan and Shalit, Uri and Sontag, David},
  year    = {2018},
  journal = {arXiv: Machine Learning},
  month   = {02},
  pages   = {},
  title   = {Learning Weighted Representations for Generalization Across Designs},
  doi     = {10.48550/arXiv.1802.08598}
}

@article{30,
  author   = {Hal R. Varian },
  title    = {Causal inference in economics and marketing},
  journal  = {Proceedings of the National Academy of Sciences},
  volume   = {113},
  number   = {27},
  pages    = {7310-7315},
  year     = {2016},
  doi      = {10.1073/pnas.1510479113},
  url      = {https://www.pnas.org/doi/abs/10.1073/pnas.1510479113},
  eprint   = {https://www.pnas.org/doi/pdf/10.1073/pnas.1510479113}
}

@article{31,
  url         = {https://doi.org/10.1515/jci-2021-0048},
  title       = {Causal inference in AI education: A primer},
  author      = {Andrew Forney and Scott Mueller},
  pages       = {141-173},
  volume      = {10},
  number      = {1},
  journal     = {Journal of Causal Inference},
  doi         = {doi:10.1515/jci-2021-0048},
  year        = {2022},
  lastchecked = {2025-02-07}
}

@article{32,
  title       = {Deep learning of causal structures in high dimensions under data limitations},
  author      = {Lagemann, Kai and Lagemann, Christian and Taschler, Bernd and Mukherjee, Sach},
  pages       = {1306-1316},
  volume      = {5},
  number      = {11},
  journal     = {Nature Machine Intelligence},
  doi         = {doi:10.1038/s42256-023-00744-z},
  year        = {2023},
  lastchecked = {2025-02-07},
  url         = {https://doi.org/10.1038/s42256-023-00744-z}
}

@article{36,
  author     = {Yao, Liuyi and Chu, Zhixuan and Li, Sheng and Li, Yaliang and Gao, Jing and Zhang, Aidong},
  title      = {A Survey on Causal Inference},
  year       = {2021},
  issue_date = {October 2021},
  publisher  = {Association for Computing Machinery},
  address    = {New York, NY, USA},
  volume     = {15},
  number     = {5},
  issn       = {1556-4681},
  url        = {https://doi.org/10.1145/3444944},
  doi        = {10.1145/3444944},
  journal    = {ACM Trans. Knowl. Discov. Data},
  month      = may,
  articleno  = {74},
  numpages   = {46}
}

@article{37,
  author     = {Huang, Qiang and Ma, Jing and Li, Jundong and Guo, Ruocheng and Sun, Huiyan and Chang, Yi},
  title      = {Modeling Interference for Individual Treatment Effect Estimation from Networked Observational Data},
  year       = {2023},
  issue_date = {April 2024},
  publisher  = {Association for Computing Machinery},
  address    = {New York, NY, USA},
  volume     = {18},
  number     = {3},
  issn       = {1556-4681},
  url        = {https://doi.org/10.1145/3628449},
  doi        = {10.1145/3628449},
  journal    = {ACM Trans. Knowl. Discov. Data},
  month      = dec,
  articleno  = {48},
  numpages   = {21}
}

@article{huang2024entire,
  title={Entire Chain Uplift Modeling with Context-Enhanced Learning for Intelligent Marketing},
  author={Yinqiu Huang and Shuli Wang and Min Gao and Xue Wei and Changhao Li and Chuan Luo and Yinhua Zhu and Xiong Xiao and Yi Luo},
  journal={Companion Proceedings of the ACM Web Conference 2024},
  year={2024},
  url={https://api.semanticscholar.org/CorpusID:267499863}
}

@ARTICLE{afzal2024multi,
  author={Afzal, Ifra and Yilmazel, Burcu and Kaleli, Cihan},
  journal={IEEE Access}, 
  title={An Approach for Multi-Context-Aware Multi-Criteria Recommender Systems Based on Deep Learning}, 
  year={2024},
  volume={12},
  number={},
  pages={99936-99948},
  doi={10.1109/ACCESS.2024.3428630}}

@inproceedings{sun2025robust,
    author = {Sun, Zexu and Han, Qiyu and Zhu, Minqin and Gong, Hao and Liu, Dugang and Ma, Chen},
    title = {Robust Uplift Modeling with Large-Scale Contexts for Real-time Marketing},
    year = {2025},
    isbn = {9798400712456},
    publisher = {Association for Computing Machinery},
    address = {New York, NY, USA},
    url = {https://doi.org/10.1145/3690624.3709293},
}

@misc{wei2024mtmt,
    title={Multi-Treatment Multi-Task Uplift Modeling for Enhancing User Growth}, 
    author={Yuxiang Wei and Zhaoxin Qiu and Yingjie Li and Yuke Sun and Xiaoling Li},
    year={2024},
    eprint={2408.12803},
    archivePrefix={arXiv},
    primaryClass={cs.LG},
    url={https://arxiv.org/abs/2408.12803}, 
}

@inproceedings{kallus2020deepmatch,
    author = {Kallus, Nathan},
    title = {DeepMatch: balancing deep covariate representations for causal inference using adversarial training},
    year = {2020},
    publisher = {JMLR.org},
    booktitle = {Proceedings of the 37th International Conference on Machine Learning},
    articleno = {470},
    numpages = {11},
    series = {ICML'20}
}

@misc{dr2020incremental,
      title={Incremental causal effects}, 
      author={Dominik Rothenhäusler and Bin Yu},
      year={2020},
      eprint={1907.13258},
      archivePrefix={arXiv},
      primaryClass={stat.ME},
      url={https://arxiv.org/abs/1907.13258}, 
}

@misc{zhang2025doubly,
      title={Doubly Robust Inference on Causal Derivative Effects for Continuous Treatments}, 
      author={Yikun Zhang and Yen-Chi Chen},
      year={2025},
      eprint={2501.06969},
      archivePrefix={arXiv},
      primaryClass={stat.ME},
      url={https://arxiv.org/abs/2501.06969}, 
}

@article{2022DESCN, 
    title={DESCN: Deep Entire Space Cross Networks for Individual Treatment Effect Estimation},
    url={http://dx.doi.org/10.1145/3534678.3539198},
    DOI={10.1145/3534678.3539198},
    booktitle={Proceedings of the 28th ACM SIGKDD Conference on Knowledge Discovery and Data Mining},
    publisher={ACM},
    author={Zhong, Kailiang and Xiao, Fengtong and Ren, Yan and Liang, Yaorong and Yao, Wenqing and Yang, Xiaofeng and Cen, Ling},
    year={2022},
    month=aug, pages={4612–4620},
    journal = {Proceedings of the 28th ACM SIGKDD Conference on Knowledge Discovery and Data Mining}, 
    volume = {4612–4620},
    series={KDD ’22}
}

@article{2008imben,
author = {Imbens, Guido and Wooldridge, Jeffrey},
year = {2008},
month = {09},
pages = {5-86},
title = {Recent Developments in the Econometrics of Program Evaluation},
volume = {47},
journal = {Journal of Economic Literature},
doi = {10.3386/w14251}
}

@article{2018chernozhukov,
    author = {Chernozhukov, Victor and Chetverikov, Denis and Demirer, Mert and Duflo, Esther and Hansen, Christian and Newey, Whitney and Robins, James},
    title = {Double/debiased machine learning for treatment and structural parameters},
    journal = {The Econometrics Journal},
    volume = {21},
    number = {1},
    pages = {C1-C68},
    year = {2018},
    month = {01},
    issn = {1368-4221},
    doi = {10.1111/ectj.12097},
    url = {https://doi.org/10.1111/ectj.12097},
    eprint = {https://academic.oup.com/ectj/article-pdf/21/1/C1/27684918/ectj00c1.pdf},
}

@inproceedings{2025qini,
author = {Goldenberg, Dmitri and Proen\c{c}a, Hugo Manuel and Livne, Amit and Moraes, Felipe and Albert, Javier and Shapira, Bracha},
title = {Converted Data is All You Need for Causal Optimization of e-Commerce Promotions},
year = {2025},
isbn = {9798400720406},
publisher = {Association for Computing Machinery},
address = {New York, NY, USA},
url = {https://doi.org/10.1145/3746252.3761573},
doi = {10.1145/3746252.3761573},
booktitle = {Proceedings of the 34th ACM International Conference on Information and Knowledge Management},
pages = {5666–5673},
numpages = {8},
location = {Seoul, Republic of Korea},
series = {CIKM '25}
}

@article{Liu2024BenchmarkingFD,
  title={Benchmarking for Deep Uplift Modeling in Online Marketing},
  author={Dugang Liu and Xing Tang and Yang Qiao and Miao Liu and Zexu Sun and Xiuqiang He and Zhong Ming},
  journal={ArXiv},
  year={2024},
  volume={abs/2406.00335},
  url={https://api.semanticscholar.org/CorpusID:270214925}
}

@article{DeepIsotonic2024,
title = {Deep Isotonic Embedding Network: A flexible Monotonic Neural Network},
journal = {Neural Networks},
volume = {171},
pages = {457-465},
year = {2024},
issn = {0893-6080},
doi = {https://doi.org/10.1016/j.neunet.2023.12.026},
url = {https://www.sciencedirect.com/science/article/pii/S0893608023007311},
author = {Jiachi Zhao and Hongwen Zhang and Yue Wang and Yiteng Zhai and Yao Yang}
}

@article{mmd2012,
author = {Gretton, Arthur and Borgwardt, Karsten M. and Rasch, Malte J. and Sch\"{o}lkopf, Bernhard and Smola, Alexander},
title = {A kernel two-sample test},
year = {2012},
issue_date = {3/1/2012},
publisher = {JMLR.org},
volume = {13},
number = {null},
issn = {1532-4435},
journal = {J. Mach. Learn. Res.},
month = mar,
pages = {723–773},
numpages = {51}
}

@article{Propensity_score2023,
title = {Propensity score methods in observational research: brief review and guide for authors},
journal = {British Journal of Anaesthesia},
volume = {131},
number = {5},
pages = {805-809},
year = {2023},
issn = {0007-0912},
doi = {https://doi.org/10.1016/j.bja.2023.06.054},
url = {https://www.sciencedirect.com/science/article/pii/S0007091223003604},
author = {Benjamin Y. Andrew and M. {Alan Brookhart} and Rupert Pearse and Karthik Raghunathan and Vijay Krishnamoorthy}
}

@article{AdversariallyBalanced2023,
  title={Adversarially Balanced Representation for Continuous Treatment Effect Estimation},
  author={Amirreza Kazemi and Martin Ester},
  journal={ArXiv},
  year={2023},
  volume={abs/2312.10570},
  url={https://api.semanticscholar.org/CorpusID:266348271}
}

@article{2026two_stage_repr_balance,
title = {A two-stage disentangled and balanced representation learning method for counterfactual regression},
journal = {Information Sciences},
volume = {730},
pages = {122886},
year = {2026},
issn = {0020-0255},
doi = {https://doi.org/10.1016/j.ins.2025.122886},
url = {https://www.sciencedirect.com/science/article/pii/S0020025525010229},
author = {Siyi Wang and Yiyan Huang and Cheuk Hang Leung and Chaoqun Wang and Qi Wu}
}

\end{document}